\pdfoutput=1

\documentclass[11pt,dvipsnames]{article}

\usepackage{EMNLP2023}

\usepackage{times}
\usepackage{latexsym}
\usepackage{amsmath}
\usepackage{tabularx}
\usepackage{multirow}
\usepackage{booktabs}
\usepackage{graphicx}
\usepackage[export]{adjustbox}
\usepackage{caption}
\usepackage{subcaption}
\usepackage{enumitem}
\usepackage{xspace}
\usepackage{scalerel,xparse}

\usepackage[T1]{fontenc}

\usepackage[utf8]{inputenc}

\usepackage{microtype}

\usepackage{inconsolata}

\definecolor{maria-color}{HTML}{7881F2}
\definecolor{yueorange}{RGB}{212,85,47}
\definecolor{yueblue}{RGB}{0,11,183}
\definecolor{yuegreen}{RGB}{0,119,87}

\title{Personalized Jargon Identification for Enhanced Interdisciplinary Communication}

\author{
    Yue Guo$^{1,*}$ \quad Joseph Chee Chang$^{2}$ \quad Maria Antoniak$^{2}$ \\
    \textbf{Erin Bransom$^{2}$ \quad Trevor Cohen$^{1}$ \quad Lucy Lu Wang$^{1,2}$ \quad  Tal August$^{2}$} \\ [1mm]
    $^{1}$University of Washington \quad
    $^{2}$Allen Institute for AI \\ [1mm]
    \texttt{\{yguo50, cohenta, lucylw\}@uw.edu}, \\
    \texttt{\{josephc, mariaa, erinbransom, tala\}@allenai.org} \\ [-1mm]
}

\begin{document}
\maketitle
\begingroup\def\thefootnote{*}\footnotetext{Work performed during internship at AI2.}\endgroup
\begin{abstract}

Scientific jargon can impede researchers when they read materials from other domains. Current methods of jargon identification mainly use corpus-level familiarity indicators (e.g., Simple Wikipedia represents plain language). However, researchers' familiarity of a term can vary greatly based on their own background. We collect a dataset of over 10K term familiarity annotations from 11 computer science researchers for terms drawn from 100 paper abstracts.
Analysis of this data reveals that jargon familiarity and information needs vary widely across annotators, even within the same sub-domain (e.g., NLP). We investigate features representing individual, sub-domain, and domain knowledge to predict individual jargon familiarity. We compare supervised and prompt-based approaches, finding that prompt-based methods including personal publications yields the highest accuracy, though zero-shot prompting provides a strong baseline. This research offers insight into features and methods to integrate personal data into scientific jargon identification.

\end{abstract}

\section{Introduction}

An important challenge to communicating knowledge across scientific domains is aligning on a shared vocabulary~\citep{Strober2006HabitsOT}. Each scientific domain has unique terminology that optimizes communication within the field but can pose a barrier to researchers in other domains~\citep{Lucy2022WordsAG, Choi2007MultidisciplinarityIA}. As science becomes more specialized, so too does its terminology~\citep{Barnett2020TheGO, readability2017}, raising the barrier of learning and collaborating across disciplines. We envision systems that can identify whether specialized terminology will be unfamiliar to an individual scholar.
\begin{figure}
    \centering
    \includegraphics[trim={11.5cm 20cm 25cm 1cm},clip,scale=0.25]{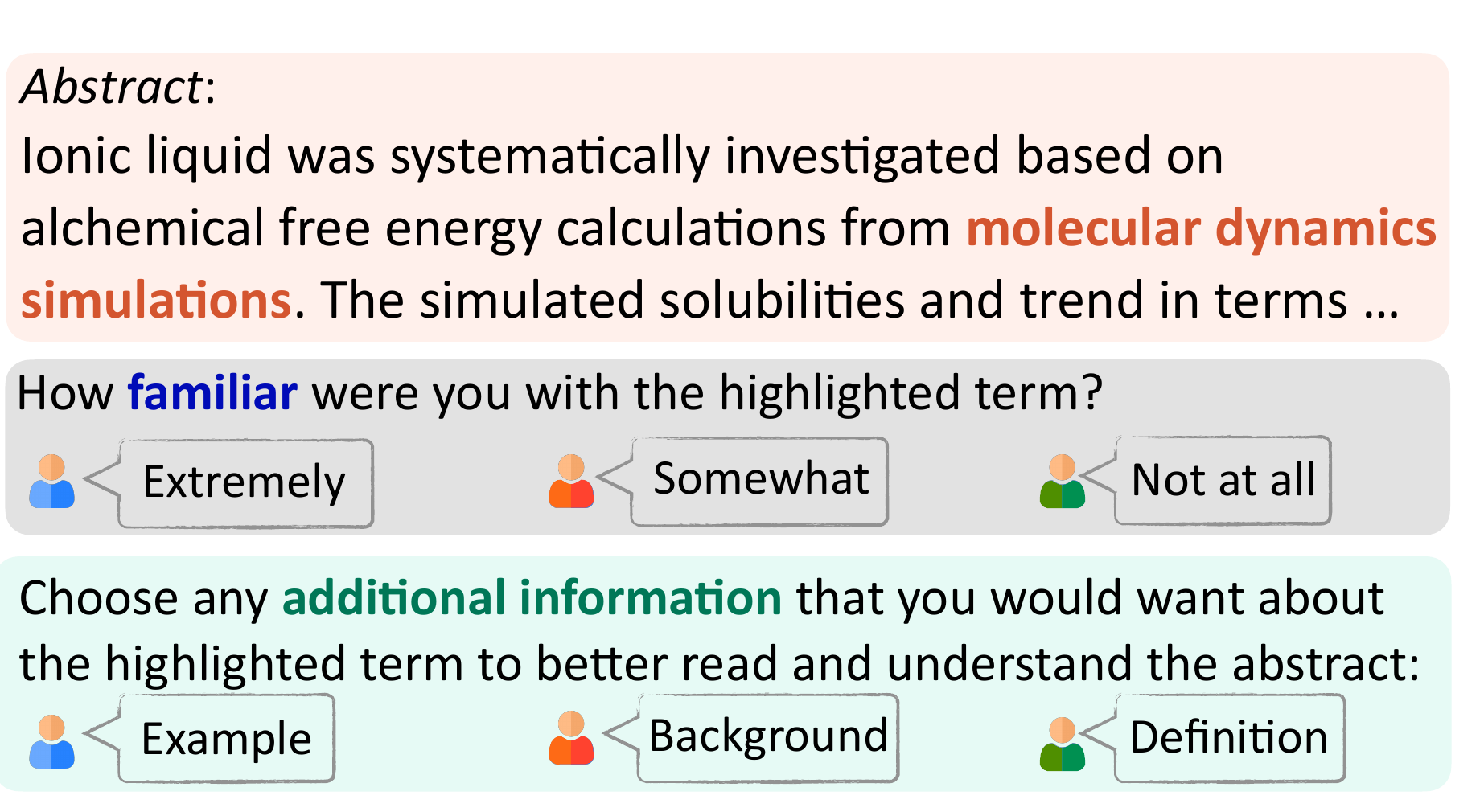}
    \caption{An annotated term from our dataset, with annotations by computer science researchers. Despite sharing a common domain, these researchers exhibit variation in their \textcolor{yueblue}{familiarity} and \textcolor{yuegreen}{additional information needs} about the \textcolor{yueorange}{term} within the abstract. Abstract from \citet{liu2014solubility}.}
    \label{fig:teaser}
\end{figure}

NLP techniques have been developed to identify and simplify scholarly jargon \citep{10.1093/applin/amt015, TanakaIshii2011WordFA, Guo2022CELLSAP, guo2021automated}, a first step in our envisioned setting. The majority of these techniques use a corpus of documents as a proxy for what a reader knows (e.g., Wikipedia contains words known to a general audience). However, an individual’s specific background knowledge also plays a role in determining their familiarity with a word \citep{Gooding2022OneSD}. For example, a theoretical computer science (CS) researcher might struggle more with jargon in a chemistry paper than in a mathematics paper, but the opposite may be true for a CS researcher in computational biology. Information on researcher's background could help determines what they know, and what they need explained.

In this paper, we investigate techniques for estimating jargon familiarity for individual researchers. We ground our investigation in the real-world setting of interdisciplinary reading: researchers reading papers in less familiar domains. We first validate our setting with an initial study around interdisciplinary reading. The results reveal a clear preference for supplementary information beyond what is provided in the abstract, especially in less familiar domains. Building on these initial findings, we propose the task of personalized scholarly jargon identification: predicting the familiarity and any associated information needs of a term for an individual researcher. 

Prior work has explored providing interactive systems to augment scientific abstracts \citep{Fok2023QlarifyBS} and provide term definitions \citep{Head2020AugmentingSP, August2022PaperPM}. We investigate a complementary task: predicting which terms and what information a researcher needs. Models that can achieve accurate, individualized predictions in this context could greatly improve interactive reading systems by focusing aids on only the most important, unfamiliar words for an individual reader, avoiding information overload \citep{Ridder2002VisibleOI, Head2020AugmentingSP}. Further, having effective predictions of term familiarity can improve writing aids by identifying terms most likely to be unfamiliar to a defined audience.

We collect a dataset of over 10k individual familiarity ratings and information needs from 11 CS researchers about terms drawn from 100 out-of-domain abstracts (example in Figure \ref{fig:teaser}). We enumerate features representing an individual's knowledge based on papers they have written and read within their domain. Using these features and our dataset, we investigate baselines for estimating term familiarity, including regression models and prompt-based approaches using large language models (LLMs). Our analysis reveals that incorporating individual-level information improves the accuracy of predicting term familiarity, though a zero-shot baseline presents a strong alternative. Our project contributes the following: 

\begin{enumerate}[itemsep=0pt, topsep=3pt, leftmargin=13pt]
    \item We define the novel task of predicting personalized jargon familiarity. We motivate our task based on initial experiments with interdisciplinary computer science researchers. 
    \item We collect a dataset of over 10k term familiarity ratings and individual information needs. Eleven CS researchers provided familiarity ratings and additional information needs for terms drawn from 100 out-of-domain abstracts.
    \item We enumerate features approximating an individual researcher's knowledge (at the domain, subdomain, and personal levels) and investigate integrating these features into supervised and prompt-based methods for term familiarity prediction. Our results show the value of domain, subdomain, and personal level features for both methods. 
\end{enumerate}

\section{Related Work}
Relevant to the current paper is research on interdisciplinary communication, scientific text simplification, and user modeling. 

\subsection{Interdisciplinary communication}

 Interdisciplinary research integrates knowledge from multiple disciplines to address a shared question ~\citep{Daniel2022ChallengesFI}. ~\citet{Choi2007MultidisciplinarityIA} surveyed interdisciplinary researchers in the health sciences, finding that a mismatch in terminology complicates efforts in communicating between disciplines. Similar findings have been reported when interviewing researchers in psychology and neuroscience ~\citep{Wudarczyk2021BringingTR}. ~\citet{Lucy2022WordsAG} found that papers that used more discipline-specific terminology (i.e., jargon), had fewer citations across disciplines, and ~\citet{Martnez2021SpecializedTR} found that papers who use more jargon are generally cited less. 

 \subsection{Scientific text simplification}
Detecting scholarly jargon is commonly done using corpus-based approaches ~\citep{TanakaIshii2011WordFA}. ~\citet{10.1093/applin/amt015} identified scholarly jargon in English by studying the frequency of words within scientific papers compared to a background corpus of general English writing. Similar methods have identified jargon in specific fields of science, including medical studies~\citep{August2022PaperPM}, and computer science papers ~\citep{Salatino2018TheCS}. Taking into account individual knowledge can also improve jargon identification ~\citep{Lee2018PersonalizingLS} and definitions ~\citep{Murthy2021TowardsPD}. ~\citet{Gooding2022OneSD} found that training models at the individual level improves complex word identification, and ~\citet{Lin2012MeasuringID} found that using social media posts written by an individual can help predict word familiarity. In work most similar to our own,~\citet{Murthy2021TowardsPD} introduced the task of defining scientific terms based on words a scientist already knows. 
In contrast to prior work, our focus on scientists provides unique opportunities to model individual knowledge. Scientists develop deep knowledge of their field by reading and writing scientific papers. We investigate techniques to leverage this information to predict personal term familiarity.

\subsection{User Modeling}
Work has sought to model the knowledge of individuals using cognitive approaches \citep{Desmarais2012ARO, Wu2021UserasGraphUM, Amith2020MiningHV}. ~\citet{Amith2017UsingPN} compared expert and consumer health knowledge on vaccines and general medical concepts with estimates of semantic distance between terms inferred from expert- and consumer-authored corpora using methods of distributional semantics. Work has also defined differences in concept knowledge between experts and lay readers for general medical concepts~\citep{Zhang2002RepresentationsOH}. User modeling approaches require either an existing set of concepts or information about how concepts are related to one another. We take a more lightweight approach by leveraging existing text representing the knowledge of an individual scientist to estimate their knowledge of new terms.

\section{Task Description}
\label{appsec:formative_study}

We conduct an initial study with 10 computer science researchers to (1) validate our intuition that term familiarity is important for interdisciplinary reading, and (2) identify what information needs scholars have for unfamiliar terms when reading across domains. 

We recruited participants from two subdomains of computer science: Natural-Language Processing (NLP) and Human-Computer Interaction (HCI). Participants were asked to read two paper abstracts, one from a closer domain: Linguistics or Psychology, and one from a more distant domain: Medicine. For each paper, we provided two abstract variants: the original author-written abstract and a generated abstract personalized to the participant's background (details can be found in App. Table~\ref{table:gpt_prompts}). After reading each abstract pair, participants identified what modifications they liked/disliked in the personalized abstract, and provided a free-text response on whether they preferred the generated abstract and why. The study was considered exempt upon University of Washington IRB [MOD00015554].

\subsection{Initial Findings}
\label{sec:initialFindings}
Of 20 responses collected from 10 researchers, 14 expressed a preference for the personalized abstract over the original, with 9 preferring the personalized abstract for the medical paper (90\%) and 5 preferring the personalized abstract for the linguistics or psychology paper (50\%). Full results can be found in App.~\ref{appsec:task_definition}. 

We categorize the participants' preferred transformations as satisfying the following information needs:
 
 \begin{itemize}[itemsep=0pt, topsep=1pt, leftmargin=10pt]
    \item \textbf{Definition}: provides key information on the term independent of any context (e.g., a specific scientific abstract). A definition answers the question, ``What is/are [term]?''
    \item \textbf{Background}: introduces information that is important for understanding the term in the context of the abstract, e.g., how the term relates to the overall problem, significance, and motivation of the paper.
    \item \textbf{Example}: offers specific instances that help illustrate the practical application or usage of the term within the abstract.
    \item \textbf{Method/Result Details}: details on the methodology and results of the paper, e.g., how models were applied or data collected.
    \item \textbf{Relevant Downstream Connections}: provides insights about how the current paper's findings relate to the reader's own research.
\end{itemize}

Participants generally indicated a preference for receiving additional information only when they were unfamiliar with a term. This suggests that an effective approach to personalizing scientific abstracts would be to first identify unfamiliar terms, then determine the applicable information need for each. In this work, we focus on the first three needs---definitions, background, and examples---as these were typically associated with researchers being unfamiliar with the abstract domain (24 out of 30 of the preferred changes were in the medical abstract). We omit method/result details from further study because these needs usually surfaced when a researcher was more familiar with a domain or particular method. While participants generally reacted positively to relevant downstream connections, these were usually hallucinated by the model; we consequently avoid targeting these as well.

\subsection{Task Definition}
Based on these initial findings, we identify the tasks of individual term familiarity prediction and information need prediction as important steps for assisting interdisciplinary reading of scientific abstracts. We formalize the first task as: given an individual researcher defined by their authored publications $R =\{r_1, r_2, ... r_m \}$ and an abstract to personalize $A$, which includes a set of terms $T =\{t_1, t_2, ... t_n \}$, our goal is to predict the subset of terms unfamiliar to $R$.

\section{Dataset}
\label{sec:evaluation_dataset}

As no pre-existing datasets exist for personalized scientific jargon identification, we construct a new dataset of terms from abstracts with human annotations of familiarity and additional information needs. We direct our focus to abstracts that are outside the individual's domain, with CS researchers as the annotators.

\subsection{Data Source}
To ensure that the out-of-domain abstracts could realistically be read by our annotators, we compile a corpus of non-CS papers often viewed by CS researchers, published after 2010, using the Semantic Scholar API \citep{Kinney2023TheSS}. We define CS researchers as anyone who has co-authored a paper categorized as `Computer Science' using the API. We take the top 500 viewed papers not categorized as CS.   
To construct a representative corpus covering diverse domains, 
we randomly draw from each domain such that the resulting numbers of abstracts across all domains are proportionally the same as those from the most-viewed non-CS papers.
This method was employed to select 100 abstracts.
For each abstract, the top-10 significant terms are identified using the OpenAI model \textit{text-davinci-003} (details and prompt in App. Table~\ref{table:gpt_prompts}). We manually review 10 abstracts and confirm that these top 10 terms align with our notion of salient terms in each abstract. We define salient terms as terms that could be provided as keyword descriptors of the paper.

\subsection{Annotation}
We recruit annotators that are: (1) earning at least a master's degree in CS, and (2) an author on at least one published work. Annotators were asked to annotate each term with the following information:
\begin{itemize}[itemsep=0pt, topsep=1pt, leftmargin=10pt]
    \item \textbf{Familiarity} on a scale of 1 (not at all familiar) to 5 (extremely familiar). Not at all familiar was defined as `you have never heard of this term.' Extremely familiar was defined as `you have a deep, comprehensive understanding of this term.'
    \item \textbf{Additional information needs} that could help annotators understand the abstract. These included definitions, background, and examples (defined in \S\ref{sec:initialFindings}). Annotators could select more than one information need. 
\end{itemize}

\noindent We recruited 11 annotators with Master's (N=4) or Doctorate (N=7) degrees in CS via UpWork, and paid each annotator \$20-30 hourly based on their degree. Each annotator reviewed terms from all sampled abstracts and answered all questions. The 11 annotators are from 17 self-defined subfields, including computer vision, machine learning, NLP, cognitive science, education, computer networks, etc. Their numbers of publications varied from 1 to 60, with a median publication count of 10. 

\subsection{Outcomes}
We define a binary term familiarity outcome measure by grouping the collected 5-point familiarity ratings into the binary classes of ``familiar'' (ratings $\geq$ 3) and ``unfamiliar'' (ratings $\leq$ 2). For terms annotated as requiring further information, we also treat the need for additional definitions, background, and/or examples as labels for a secondary classification task.

\subsection{Analysis}
Our dataset includes 956 terms sourced from 100 abstracts from 22 domains.\footnote{GPT 3.5 identified <10 terms from some abstracts. Upon inspecting these abstracts, the authors agreed that there were fewer than 10 important terms to list.} We collected a total of 10,571 familiarity ratings and responses for additional information needs from the 11 annotators. 

\begin{figure}[t!]
    \centering
    \includegraphics[trim={23cm 13cm 25cm 9cm},clip,scale=0.4]{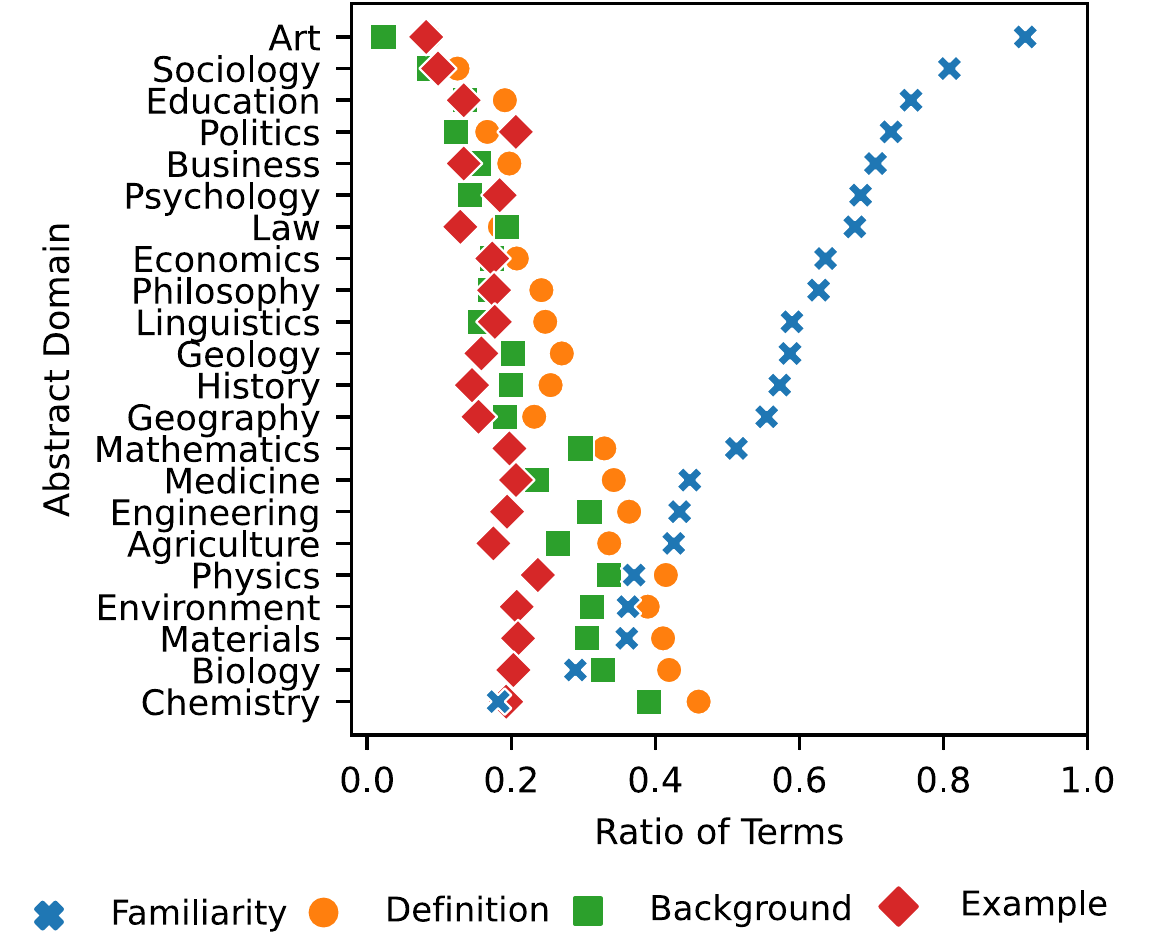}
    \caption{Mean familiarity and additional information needs (definition, background, and example) across abstract domains. The ratio of terms shows how many terms in the abstract domain are familiar, and require definitions, background, and examples.}
    \label{fig:familiarity_additional_info_by_domain}
\end{figure}

\paragraph{Data check}
To ensure that annotator familiarity ratings were consistent, we conducted a data check for all annotators. We selected 10 entities from each annotator, 5 rated as familiar and 5 rated as unfamiliar. For each entity, we asked annotators to provide a definition of the entity without looking up any information. If they could not define the term, we instructed them to write `N/A'. Annotators were generally consistent with their initial scores, with 81\% of their responses matching their initial ratings (i.e., if they were familiar, they wrote a correct definition). In cases where initial scores did not align with the data check, annotators generally wrote definitions given by context in the abstract. 

\paragraph{Domain-specific variations}
Figure~\ref{fig:familiarity_additional_info_by_domain} illustrates the differences in familiarity and additional information needs across abstract domains. Generally, we see that annotators most often select high familiarity for Art while Chemistry receives the lowest familiarity ratings. This represents a general trend of annotators selecting lower familiarity for technical sciences than for humanities or social sciences. The same trend holds for additional information needs; annotators request the most definitions and background for terms in Chemistry, and the least in Art. Interestingly, though, annotators in general prefer a baseline rate of examples that does not vary greatly across domains.

\paragraph{Individual-specific variations}
19\% of the terms receive uniform familiarity ratings from all annotators.
Consensus in familiarity is most frequently found within terms related to Philosophy and Business. Conversely, terms associated with Chemistry and Biology are consistently marked as unfamiliar. Notably, there is an even split in familiarity for 15\% of the terms---half deemed familiar and half not---with Mathematics and Engineering being the predominant fields for such terms. This diversity highlights the nuances of individual knowledge backgrounds even within the common discipline of CS. 

\section{Predicting Term Familiarity}

Given our dataset of term familiarity ratings and annotator backgrounds, we investigate the effectiveness of a set of features and methods for predicting individual term familiarity.

\subsection{Features}
Past work has explored using readability measures, frequency statistics, and embeddings to predict term familiarity at the population-level \citep{Rakedzon2017AutomaticJI, August2022PaperPM, Lucy2022WordsAG}. We adapt the following features for predicting individual-level term familiarity: 
\begin{itemize}[itemsep=0pt, topsep=1pt, leftmargin=10pt]
    \item \textbf{Frequency}: The number of times a term appears in researcher's publications $R$. 
    \item \textbf{Specificity}: Quantifies the term's uniqueness to a corpus \citep{zhang2017community}, computed as the log probability ratio:
    \vspace{-2mm}
    \begin{equation*}
    S_c(t) = log\frac{P_c(t)}{P_C(t)}
    \end{equation*}
    In our case, $c$ corresponds to the target abstract $A$ and $C$ to the researcher's publications $R$.
    \item \textbf{Embedding similarity}: Leverages SPECTER \cite{specter2020cohan}, a citation-based transformer model encoding semantic document similarity. We calculate the minimum Euclidean distance between the target abstract $A$'s embedding and any of the author's publications $R = \{r_1, r_2,...\}$'s embeddings as its SPECTER similarity. 
\end{itemize}

\noindent For each feature, we start by defining different granularities of researcher's publications $R$, representing domain, subdomain, and individual-level information. The following data is extracted from the Semantic Scholar API \citep{Kinney2023TheSS}.

\begin{itemize}[itemsep=0pt, topsep=1pt, leftmargin=10pt]
    \item \textbf{Domain}: 10k randomly sampled CS papers from 2015-2022.
    \item \textbf{Subdomain}: 10k randomly sampled papers from each annotator's self-defined subdomain corpus from CS papers from 2015-2022. Subdomain corpora are defined based on venues associated with a given subdomain.
    \item \textbf{Individual}: Represented by an individual's publications. If the individual's number of publications is less than the necessary number for training in \S\ref{sec:models}, the remaining quantity is supplemented by a random selection from the cited references within those publications.
\end{itemize}

\noindent In addition to these granular features, we include the following general-purpose measures of readability and metadata.

\begin{itemize}[itemsep=0pt, topsep=1pt, leftmargin=10pt]
    \item \textbf{Readability}: We utilize the Flesch-Kincaid (F-K) \cite{flesch2007flesch} readability score and the GPT-2 perplexity score \cite{martinc2021supervised}. F-K score is computed at the passage level; all terms from an abstract are assigned the same F-K score. 
    \item \textbf{Metadata}: We include the target paper domain, as well as the individual's attributes, including year of their first published paper, total number of published papers, and their average citation count for published papers.
\end{itemize}

\subsection{Models}
\label{sec:models}

We explore two modeling approaches: supervised and prompt-based:
\paragraph{Lasso regression}
We adopt a Logistic regression model with L1 regulation to integrate features across various levels of granularity and determine feature importance. A binary label determination is made using a threshold value of 0.5. We train one model per annotator. The model training is divided into two distinct settings: 
\begin{itemize}[noitemsep, topsep=1pt, leftmargin=10pt]
    \item \textit{Individual model}: the model is trained using data derived from terms annotated by the specific annotator alone.
    \item \textit{Mixed model}: the model is trained on familiarity ratings from all other annotators (i.e., leave-one-annotator-out testing). To maintain the same sample size of training data as each individual Lasso model, for each annotator we randomly select the same number of training data points as was used to train their individual model. 
\end{itemize}

\paragraph{Prompt-based LLMs}
We design prompts to complete a binary classification task of familiarity based on identified terms and the target abstract, using the GPT-4 model from OpenAI.
To explore different strategies, we use: 
\begin{itemize}[noitemsep, topsep=1pt, leftmargin=10pt]
    \item \textit{Baseline}: providing only the term and abstract containing the term.
    \item \textit{Metadata}: providing annotator metadata along with the baseline prompt.
    \item \textit{Context-enhanced learning}: providing publications from the annotator's domain, subdomain, and individual level separately.
    This configuration requires less data collection efforts as no labeled data is needed.  
    \item \textit{Few-shot learning}: providing examples of term, abstract, and familiarity ratings. The ratings are drawn from the three levels of granularity: ratings from other annotators with no overlap in subdomain (domain), ratings from other annotators within the same subdomain (subdomain), and ratings from the individual annotator (individual) 
\end{itemize}
Five examples are given for context-enhanced learning (e.g., abstracts from five publications authored by the annotator) and few-shot learning (e.g., five term, abstract, and familiarity labels from the annotator at the abstract level.

\subsection{Evaluation}
We split the entities randomly into an 80/20 train/test set for each annotator. All the models are then evaluated on the same test set for each annotator. The F1 score is reported to measure the model's performance on the classification task. To determine the critical features in a Lasso model, we count the features with non-zero value coefficients. A higher frequency denotes greater and more consistent influence of the feature on prediction.

\section{Results}

\paragraph{RQ1. How do supervised and prompt-based methods perform?}
\begin{figure}[t!]
    \centering
    \includegraphics[trim={24cm 12.5cm 24cm 12cm},clip,scale=0.36]{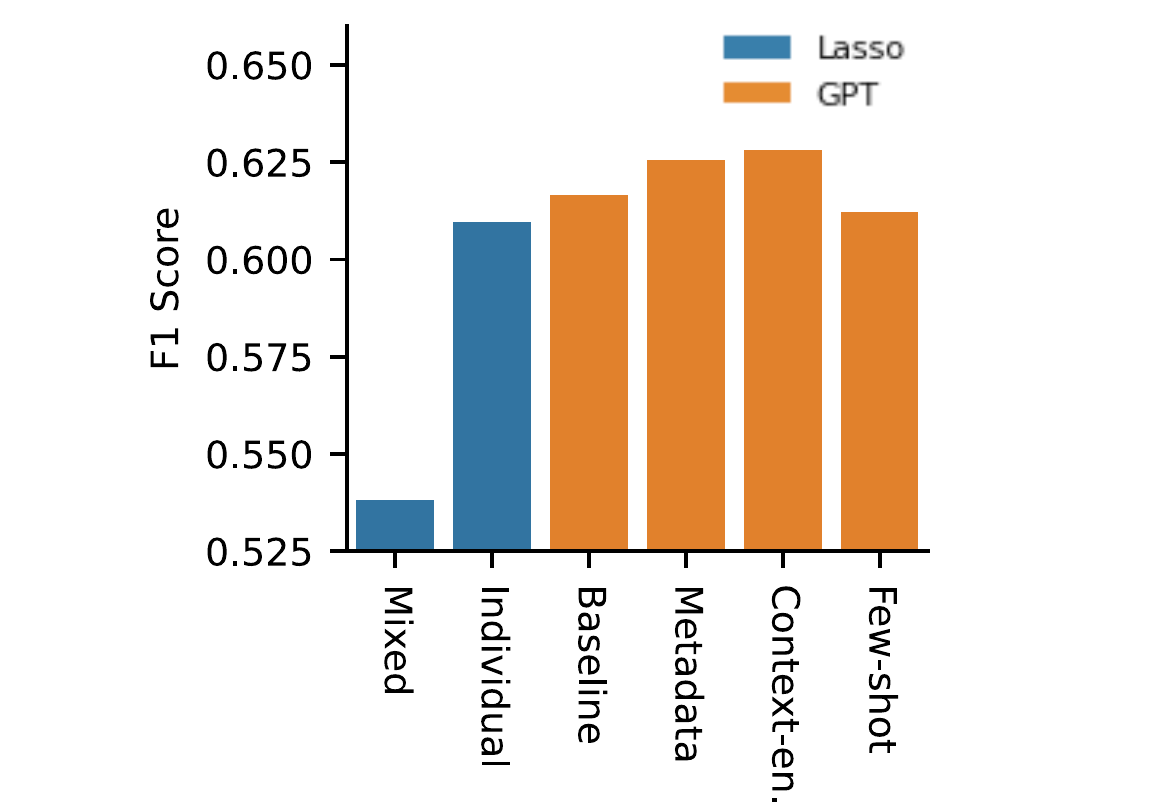}
    \caption{Model performance on term familiarity prediction for mixed and individual Lasso regression (blue) and GPT-4 with four prompt settings (orange).
    }
    \label{fig:familiarity_overall}
\end{figure}
In Figure~\ref{fig:familiarity_overall}, we present a comparison of F1 score across the highest performing predictive models. In the Lasso models, features from multiple levels of granularity are integrated. As expected, the individual Lasso model shows superior performance compared to the mixed Lasso, due to the fact that the training data for individual Lasso was from the same annotator. Despite being provided less data, GPT-4 outperforms Lasso in all settings. Metadata and context-enhanced learning perform better than few-shot learning, suggesting that the LM can infer some background knowledge of annotators from metadata or free text.

\paragraph{RQ2. What features and granularity level influence performance?}
\begin{figure}[t!]
    \centering
    \includegraphics[trim={21cm 13cm 28cm 13cm},clip,scale=0.41]{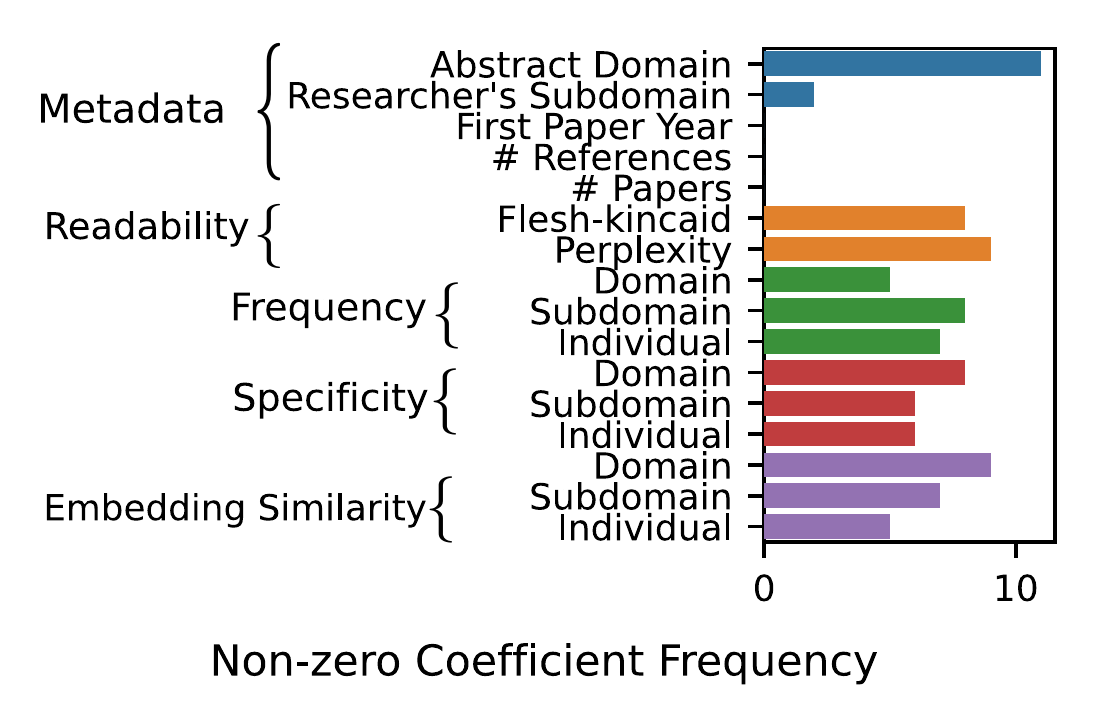}
    \caption{Frequency of non-zero Coefficients in individual Lasso models across researchers. The Lasso penalty minimizes less critical coefficients to zero. Features with higher frequencies of non-zero values in individual models are consistently identified as important.}
    \label{fig:familiarity_lasso_coefficient}
\end{figure}

\begin{figure}[t!]
    \centering
    \includegraphics[trim={21cm 10cm 22cm 9cm},clip,scale=0.3]{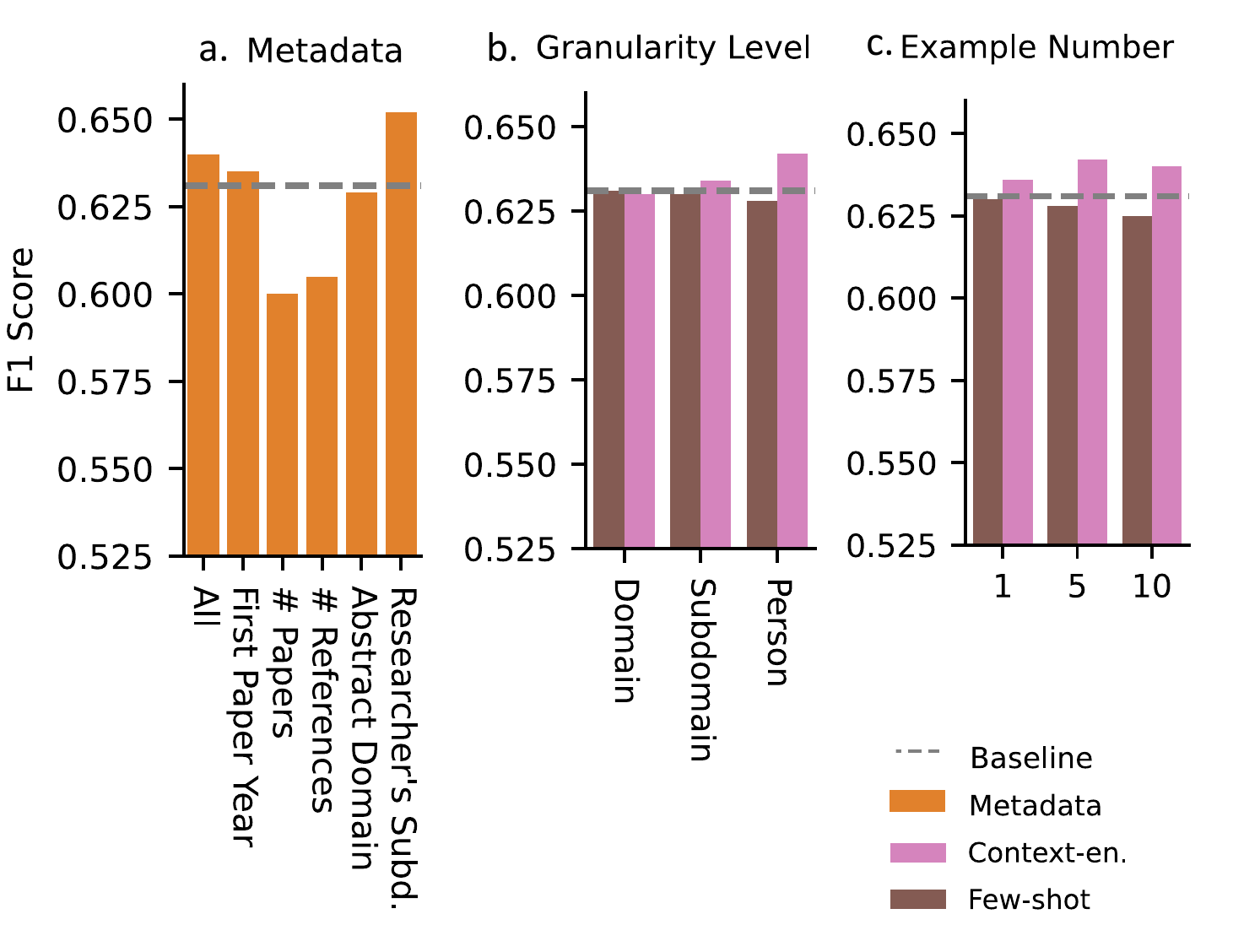}
    \caption{GPT-4 model performance: (a) with metadata, (b) at diverse granularity levels with a fixed example number of five, and (c) with varying example numbers at the personal granularity level.}
    \label{fig:familiarity_gpt_feature}
\end{figure}

Figure~\ref{fig:familiarity_lasso_coefficient} reveals the non-zero coefficients of the Lasso models. The domain of the target abstract is consistently identified as a significant feature by all individual models across annotators, indicating that an abstract's content is important for determining term familiarity, independent of individual features. Key aspects like word specificity and embedding similarity at the domain level underscore the relevance of a researcher's domain in their familiarity with jargon, aligning with previous population-level familiarity research \citep{Li2020KeywordsGuidedAS}. However, the importance of individual-level features such as frequency, specificity, and embedding similarity also emerge, highlighting the dual influence of both domain-specific and individual-specific factors in accurately predicting term familiarity. Regarding GPT-4 model performance (shown in Figure~\ref{fig:familiarity_gpt_feature}), prompting with researcher subdomain information seems to be the most effective strategy, suggesting that subdomain information (e.g., that a researcher is in NLP) is useful beyond a researcher's broad field (e.g., CS) . 

\paragraph{RQ3. How does the amount of labeled data affect task performance?}
\begin{figure}
    \centering
    \includegraphics[scale=0.72]{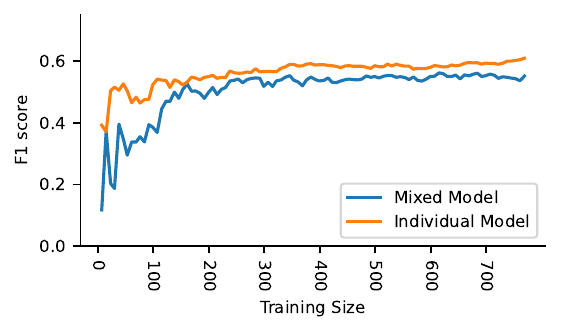}
    \caption{Performance of the Lasso model with varying training sizes. }
    \label{fig:familiarity_lasso_number}
\end{figure}

To explore this, we incrementally increase the training data for Lasso models. As indicated in Figure~\ref{fig:familiarity_lasso_number}, both mixed and individual models reach a performance plateau after the inclusion of 200 terms. Figure~\ref{fig:familiarity_gpt_feature}(c) shows that for GPT models, performance deteriorates with more examples in few-shot learning, potentially due to cross-domain data confusing the LLM's ability to predict individual level familiarity. For context-enhanced learning, five examples is more effective than ten, indicating that for LLMs, having too many examples is not necessarily beneficial.
The performance across all granularity levels with a range of example numbers can be seen in Appendix Figure~\ref{appfig:familiarity_gpt_number_granularity}.

\section{Discussion}
This paper introduces the novel task of scholarly jargon detection for the individual researcher. This task is motivated by the needs of interdisciplinary research, where researchers may need to read abstracts in domains outside their own. By focusing on domain- and personal-level data in term familiarity prediction, our research provides an analysis of the type and granularity of features, methods, and modeling strategies useful in the task of personalized jargon identification.

In pursuit of this task, we collect a dataset of over 10K term familiarity and information need annotations from 11 CS researchers for terms from 100 out-of-domain abstracts. Our dataset reveals significant variations in term familiarity among CS researchers, underscoring the diversity of individual knowledge. Additionally, the data shows that term familiarity is not always associated with the information needs (e.g., a researcher might desire an example even if they are familiar with a term), highlighting that information needs might need to be addressed differently than term familiarity.

Our findings show that unlabeled individual-level data can improve the prediction of term familiarity. This is particularly evident in prompt-based methodologies, where the use of an individual's publications as unlabeled data proved more beneficial than term familiarity labels from that individual. This might be because labeled examples did not provide enough relevant information to a model (e.g., a single term) compared to the potentially rich information found in a researcher's publications.  

Domain and subdomain level data (e.g., the subdomain of the researcher) also benefited prompt-based and regression models. This suggests that aggregated data can be effective substitutes for individual labels, a promising direction given the cost of collecting researcher-labeled term familiarity. Furthermore, for researchers with a small publication history (and therefore a small amount of unlabeled individual-level data), domain and subdomain data can provide valuable information for predicting jargon familiarity. 

Our methods provide an exciting first step to support researchers reading abstracts outside of their domain \citep{Wudarczyk2021BringingTR}. Our dataset and findings show that models can identify unfamiliar terms for individual researchers with some accuracy. Combined with text simplification techniques \citep{Srikanth2020ElaborativeSC}, models that can predict an individual researcher's familiarity with a term and any subsequent information needs could assist researchers by rewriting an abstract tuned to that particular researcher, thereby minimizing the barriers of interdisciplinary communication.

\section{Conclusion}

This paper introduces the novel task of scientific jargon detection for the individual researcher. We collect a dataset of over 10k term familiarity annotations from computer science researchers and investigate supervised and prompt-based methods to predict term familiarity. We find that leveraging a researcher's publication history, self-reported subdomain, and general domain information can improve term familiarity prediction. Our results provide insight on integrating an individual's knowledge into scientific jargon detection.

\section*{Limitations}

We focus on computer science researchers for both selecting annotators and relevant abstracts in other fields. This might limit our ability to generalize to other domains or researchers. Because of the cost of collecting annotations, our dataset is also relatively small, which might further limit its generalizability. Our goal with the dataset and analysis is to show the potential of modeling individual term familiairty and information needs. We are excited about future work expanding this goal into new domains.

\section*{Ethics Statement}

Some of the methods in the paper include personal data (e.g., publication record, labeled terms), which might pose a privacy risk for some researchers. Systems identifying term familiarity and information needs must keep any personal data stored locally and allow researchers to remove or view their data at any time. Focusing on computer science researchers as a first step for predicting term familiarity might also allow computer science researchers to more effectively read outside their discipline, but not researchers in other domains reading within computer science. While encouraging interdisciplinary reading can improve two-way communication, it is also important to consider the voices of researchers in other domains.

\bibliographystyle{acl_natbib}
\bibliography{custom}

\clearpage
\appendix
\section{Appendix}
\label{sec:appendix}

\subsection{Initial Task Definition}
\label{appsec:task_definition}
For our initial study, we focus on reading full scientific abstracts because we wanted to understand if term familiarity was an important part of our broader envisioned setting of researchers reading abstracts outside of their domain. Further, our goal in providing a naive personalized abstract was to probe what transformations are feasible with current models that researchers respond positively to. 

One abstract was from the medical domain, a domain not within any participant's dominant expertise, and one abstract was from a domain related to but distinct from a participant's specialization (e.g., psychology, linguistics). Personalized abstracts were generated using \textit{text-davinci-003} model from OpenAI \citep{OpenAI2023GPT4TR, Brown2020LanguageMA}. As a first step in including a individual background knowledge, the participant's publications were added in the prompt and the model was instructed to personalize the abstract to the given reader.

\begin{figure}[t!]
    \centering
    \includegraphics[trim={21.5cm 2.5cm 21.5cm 3cm},clip,scale=0.31]{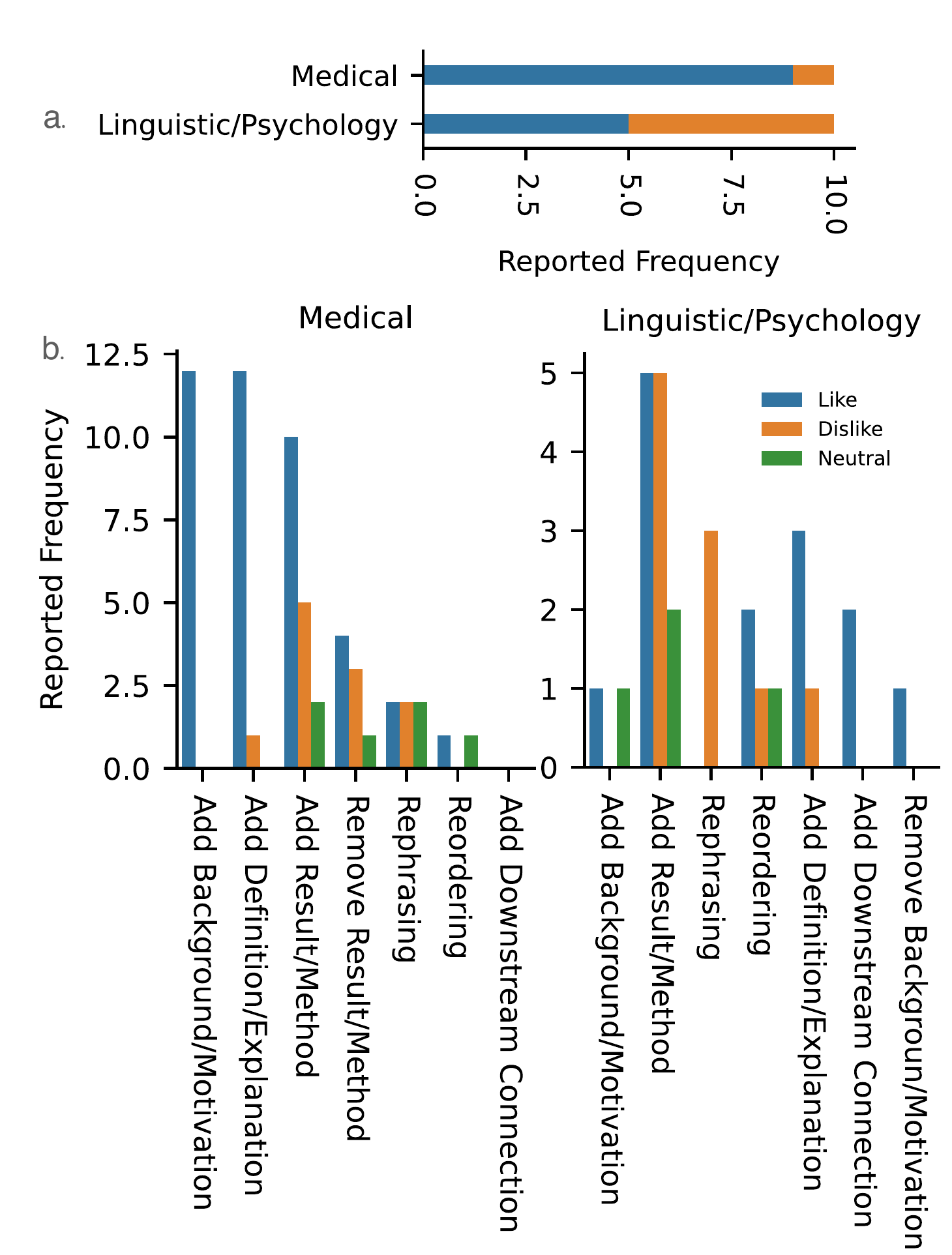}
    \caption{Formative study results: domain-specific attitudes towards (a) personalized abstracts and (b) transformations of personalized abstracts. Medical abstracts are perceived as distant from the annotators' background, whereas linguistic or psychology abstracts are perceived as closer.}
    \label{fig:formative_study}
\end{figure}

\subsection{GPT Prompts}
\label{appsec:appendix_gpt_prompts}
\begin{table*}[]
\small
\begin{tabular}{@{}m{1.4cm}m{9cm}m{1.3cm}m{1.3cm}m{1.3cm}@{}}
\toprule
Task & Prompt  & Model & Max length & Temp. \\
\midrule
Abstract personalization & You are tasked with the role as a scientific writer to generate personalized abstract for individual reader. To do this effectively, consider including relevant background information or motivations related to the subject matter, provide necessary definitions or explanations to elucidate complex concepts, and incorporate significant methodological or result-oriented details. However, ensure that any additional information included is directly relevant and can be traced back to the provided content.  The paper needed to be personalized is:\{ \};  The reader's publications are: \{ \};  The reader's references are:\{\};  The personalized abstract is: . & text-davinci-003 & 500        & 0           \\
\midrule
Top-10 significant terms & Please review the following scientific paper abstract. Your task is to identify all scientific-related word/phrases within the text and then rank these word/phrases in descending order based on their significance within the abstract itself. Retain the first 10 word/phrases:.   & text-davinci-003 & 100        & 0 \\
\midrule
Familiarity classification    & Your job is to estimate how much the reader knows about an entity. You will be provided with the entity, the abstract where the entity come from, and related data about either the reader or the abstract. Entity: \{\} Abstract:\{\} Related Data:\{\} Here's how to gauge the reader's familiarity: - 0: The reader knows this subject well and can describe it to others. - 1: The reader has either encountered this subject before but knows little about it, or has never come across it at all. Based on the information provied, determine the familiarity score, either 0 or 1:    & gpt-4            & 100        & 0 \\
\midrule
Definition needs classification & Your job is to estimate whether the reader might need additional definition to fully grasp the entities mentioned in a given abstract. You will be provided with the entity, the abstract where the entity come from, and related data about either the reader or the abstract. Definition of definition/explanation: provides key information on the term independent of any context (e.g., a specific scientific abstract). A definition answers the question, \"What is/are [term]?\". Entity: \{\} Abstract:\{\} Rel:\{\}. Provide the estimation whether additional information is needed in a list in the order of the entity. The estimation should be either 0(no) or 1(yes). No need to mention the entity: & gpt-4            & 100        & 0 \\
\midrule
Background needs classification & Your job is to estimate whether the reader might need additional background to fully grasp the entities mentioned in a given abstract. You will be provided with the entity, the abstract where the entity come from, and related data about either the reader or the abstract. Definition of background/motivation: introduces information that is important for understanding the term in the context of the abstract. Background can provide information about how the term relates to overall problem, significance, and motivation of the abstract. Entity: \{\} Abstract:\{\} Rel:\{\}. Provide the estimation whether additional information is needed in a list in the order of the entity. The estimation should be either 0(no) or 1(yes). No need to mention the entity: & gpt-4            & 100        & 0 \\
\midrule
Example needs classification & Your job is to estimate whether the reader might need additional example to fully grasp the entities mentioned in a given abstract. You will be provided with the entity, the abstract where the entity come from, and related data about either the reader or the abstract. Definition of example: offers specific instances that help illustrate the practical application or usage of the term within the abstract. Entity: \{\} Abstract:\{\} Rel:\{\}. Provide the estimation whether additional information is needed in a list in the order of the entity. The estimation should be either 0(no) or 1(yes). No need to mention the entity: & gpt-4            & 100        & 0 \\
\bottomrule
\end{tabular}
\caption{GPT-4 prompts and configurations.}
\label{table:gpt_prompts}
\end{table*}

All prompts used for GPT-4 experiments are shown in Table~\ref{table:gpt_prompts}.

\subsection{GPT-4 Results}

\begin{figure}[!t]
    \centering
    \includegraphics[trim={22cm 12cm 29cm 4cm},clip,scale=0.45]{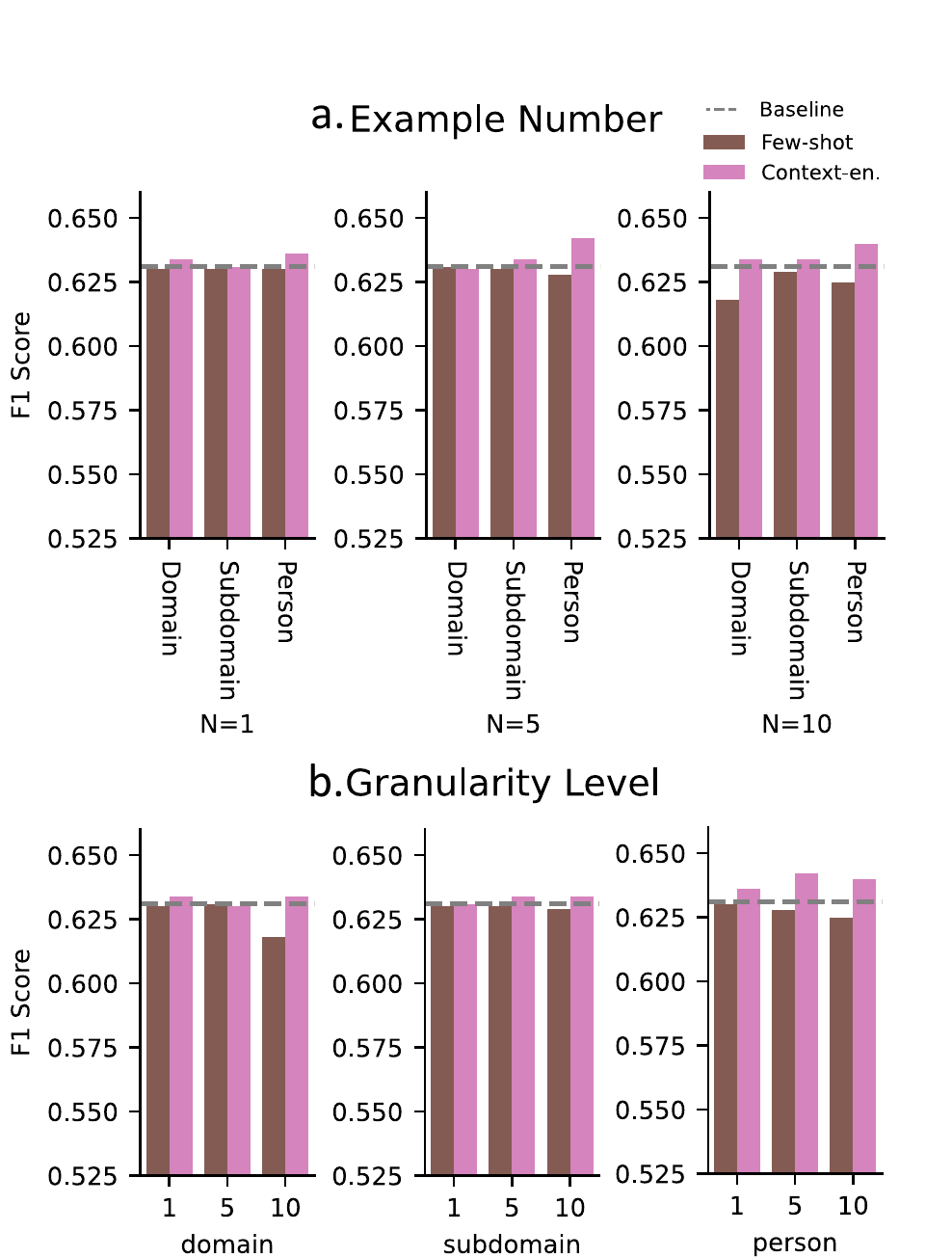}
    \caption{GPT-4 performance across all granularity levels with a range of example numbers.}
    \label{appfig:familiarity_gpt_number_granularity}
\end{figure}

GPT-4 modeling results are shown in Figure~\ref{appfig:familiarity_gpt_number_granularity}. We vary context-enhanced and ICL example numbers at N=1, 5, and 10, and the granularity of input information by domain, subdomain, and person/individual levels.

\end{document}